\documentclass[runningheads]{llncs}

\usepackage{xcolor}

\usepackage{amsmath,amssymb}

\usepackage[inline]{enumitem}

\usepackage{algorithm}

\usepackage{algpseudocode}

\usepackage{subcaption}

\usepackage{tikz}
\usetikzlibrary{arrows.meta}

\usepackage{fancybox}

\usepackage{standalone}
\usepackage{placeins}
\usepackage{graphicx}

\usepackage{csquotes}

\usepackage{hyperref}
\usepackage[
    backend=bibtex,
    natbib=true,
    sorting=nyvt,
    doi=false,isbn=false,url=false
]{biblatex}
\begin{document}
\title{Investigating the Compositional Structure Of Deep Neural Networks}
\author{Francesco Craighero$^1$ \and
Fabrizio Angaroni$^1$ \and
Alex Graudenzi$^{2,\dagger,*}$ \and
Fabio Stella$^{1,\dagger}$ \and
Marco Antoniotti$^{1,\dagger}$}
\authorrunning{F. Craighero et al.}
\institute{$^1$ Department of Informatics, Systems and Communication, \\ University of Milan-Bicocca, Milan, Italy \\
$^2$ Institute of Molecular Bioimaging and Physiology, \\ Consiglio Nazionale delle Ricerche (IBFM-CNR), Segrate, Italy \\
$^*$ corresponding author: alex.graudenzi@unimib.it \\
$\dagger$ co-senior authors
}
\maketitle              %
\begin{abstract}

The current understanding of deep neural networks can only partially explain how input structure, network parameters and optimization algorithms jointly contribute to achieve the strong generalization power that is typically observed in many real-world applications. 
In order to improve the comprehension and interpretability of deep neural networks, we here introduce a novel theoretical framework based on the compositional structure of piecewise linear activation functions. By defining a direct acyclic graph representing the composition of activation patterns through the network layers, it is possible to characterize the instances of the input data with respect to both the predicted label and the specific (linear) transformation used to perform predictions. Preliminary tests on the MNIST dataset show that our method can group input instances with regard to their similarity in the internal representation of the neural network, providing an intuitive measure of input complexity. %

\end{abstract}

\section{Introduction}
Despite the extremely successful application of Deep Neural Networks (DNNs) to a broad range of distinct domains, many efforts are ongoing both to deepen their understanding and improve their interpretability \cite{ gilpin_explaining_2018, ross_right_2017}. This is particularly relevant when attempting to explain their generalization performances, which are typically achieved due to over-parameterized models \cite{arpit_closer_2017, zhang_understanding_2016}. 

To this end, many works focus on the study of the \emph{expressivity} of DNNs, i.e., how their architectural properties such as, e.g., depth or width, affect the performances \cite{arpit_closer_2017, hanin_complexity_2019,hanin_deep_2019,montufar_number_2014,pascanu_number_2014,raghu_expressive_2017, serra_empirical_2019}. These works usually analyze DNNs with piecewise-linear (PWL) activation functions, such as Rectified Linear Units (ReLUs), which allow to simplify the mathematical analysis of the feature space.

In particular, given a standard multinomial classification problem, it is possible to study how a given input dataset is 
processed in the internal representation of a ReLU DNN by analyzing the \emph{activation patterns}, i.e., the sets of neurons that are active/inactive for each instance of the dataset, in each layer of the network. %

Each activation pattern uniquely defines a layer-specific activation region, i.e., the region of the input space which leads to the activation of the same pattern \cite{hanin_deep_2019}; clearly, one or more instances can be mapped on the same activation region. 
Each instance will be then characterized by a specific trajectory through activation patterns in successive layers, as a result of the composition of multiple ReLUs. Accordingly, each instance will be mapped onto distinct activation regions in each layer.
By analysing how different instances are characterized by common activation patterns and regions, it is possible to investigate how the input space is \emph{folded} for any given dataset.

In particular, the so-called compositional structure \cite{montufar_number_2014} of the activation patterns can then be exploited to \emph{interpret the elaboration} of the input data by a DNN, i.e., \enquote{to understand how data are represented and transformed throughout the network} \cite{gilpin_explaining_2018}. 
This structure can be translated into an \emph{Activation Pattern Direct acyclic graph} (APD), which we formally define in the following sections and that may represent a powerful instrument to evaluate the expressivity of a DNN with respect to a specific dataset.
 
Accordingly, by analyzing how many distinct instances are mapped on shared sub-portions of the APD, i.e., belong to the same activation regions, it is possible to provide an intuitive measure of the input complexity, which can be then related to  classification accuracy. 
We remark that the analysis of the relation between input data and the representation of DNNs is an active area of research in the sphere of explainable AI and covers topics such as, e.g., importance sampling \cite{chang_active_2017,kumar_self-paced_2010, katharopoulos_not_2018}.

In this work, we propose a new framework to quantitatively analyze the compositional structure of DNNs and, in particular:  
\begin{enumerate}
    \item we introduce and define the concept of Activation Pattern Direct acyclic graph (APD);
    \item we describe a lightweight algorithm to cluster the instances of a dataset on the basis of their mapping on the APD;
    \item we present an empirical analysis of the MNIST dataset \cite{lecun_mnist_2010}, in which we show that the proposed clustering method on the APD could be employed as an importance sampling algorithm.
\end{enumerate}

\section{Related works}
\label{sec:sota}

The literature devoted to the study of \emph{network expressivity} of ReLU DNNs is vast.
Three topics are particularly relevant for the current work, namely: $(i)$ the estimation of the upper-bound of the number of linear regions \cite{montufar_number_2014,pascanu_number_2014,serra_empirical_2019}; $(ii)$ the analysis of the linear regions through input trajectories \cite{raghu_expressive_2017}; $(iii)$ the analysis of other linear regions properties, such as their size or their average number \cite{hanin_complexity_2019,hanin_deep_2019}.

With respect to sample analysis, a variety of works demonstrates how sampling instances by importance during training can improve learning. Again, to limit the scope of our investigation, we can distinguish four different sampling strategies: $(i)$ curriculum learning \cite{bengio_curriculum_2009}, according to which it is preferable to start learning from easier to harder instances,  also implemented in self-paced learning \cite{kumar_self-paced_2010}; $(ii)$ selecting only the hardest instances, e.g., the ones that induce the greater change in the parameters \cite{katharopoulos_not_2018}; $(iii)$ meta-learning, i.e., \enquote{learning to learn} \cite{fan_learning_2017}; $(iv)$ favoring uncertain instances  \cite{chang_active_2017, toneva_empirical_2018}.
Given these premises, a first major challenge is the estimation of instance hardness/complexity. Accordingly, the choice of the right sampling strategy is essential and depends both on the task and on data type, e.g., (simple, noisy, \dots). 

In \cite{toneva_empirical_2018}, the authors analyze the learning process by measuring the so-called \emph{forgetting events} (defined formally in Def. \ref{def:forg_event}). An instance is called \emph{unforgettable} when no forgetting event occurs during training, otherwise it is called \emph{forgettable}. The authors show that training a new model without unforgettable samples does not affect the accuracy. 
Similarly, two further works show how to build an ensemble of DNNs by iteratively training a new network on a reduced version of the dataset. In \cite{ross_right_2017} the authors iteratively mask the features that display the greatest input gradient. As a result, they define multiple models that make predictions based on \enquote{qualitatively different reasons}, mainly to achieve greater explainability. 
In \cite{tao_deep_2019}, the authors train each new network on a reduced version of the dataset, where \enquote{good} instances of the previous network are removed (\enquote{good} inputs are the ones with hidden features belonging to mostly correctly classified instances). To define the hidden features, the authors first cluster each hidden layer with $k$-means, and then characterize each instance with respect to the clusters of each layer to which it belongs. 
We here propose a similar approach, in which each instance is characterized by the path of linear regions (activation patterns) in each layer to which it belongs.

\section{Methods}

In this section we will formally define the Activation Pattern DAG (APD) and present a novel algorithm to cluster input instances on the basis of their mapping on the APD. In the following  definitions, we will employ the notation used in \cite{montufar_number_2014}, while we refer to \cite{hanin_deep_2019} for an extensive formal description of activation patterns and activation regions.

\subsection{Basic Definitions}

Let $\mathcal N_\theta(x_0)$ be a \emph{feedforward neural network} (FNN) with input $x_0 \in \mathbb R^{n_0}$ and trainable parameters $\theta$. Each layer $h_l$, for $l \in 1,\dots,L$, is represented as a vector of dimension $n_l$, i.e.,  $h_l = [ h_{l,1}, \dots, h_{l,n_l} ]^T$, where each component $h_{l,i}$ (i.e., a neuron or unit) is the composition of a linear preactivation function $f_{l,i}$ and a nonlinear activation function $g_{l,i}$, i.e. $h_{l,i} = g_{l,i} \circ f_{l,i}$.

Let $x_l$ be the output of the $l$-th layer for $l=1,\dots, L$ and the input of the network for $l=0$, then, we define $f_{l,i}(x_{l-1}) = W_{l} x_{l-1} + b_{l,i}$, where both $W_l \in \mathbb R^{n_{l-1}}$  and $b_{l,i} \in \mathbb R$ belong to the trainable parameters $\theta$. Regarding activation functions, in this paper we will focus on piecewise linear activation functions. Thus, for the sake of simplicity, we define $g_{l,i}$ as a ReLU activation function, i.e., $g_{l,i}(x) = \max\{0, x\}$. When clear from the context, we will omit the second index of $f_{l,i}$ and $g_{l,i}$ to refer to the vector composed by all of them.

Finally, we can represent the FNN $\mathcal N_\theta$ as a function $\mathcal N_\theta: \mathbb R^{n_0} \rightarrow \mathbb R^{out}$  that can be decomposed as

\begin{equation}
\mathcal N_\theta(x) = f_{out} \circ h_L \circ \dots \circ h_1(x), 
\end{equation} 
where $f_{out}$ is the output layer (e.g., softmax, sigmoid, \dots).

\subsection{From activation patterns to the APD}

Given a FNN $\mathcal N_\theta$ and a dataset $\mathcal D$, we define the \emph{activation pattern} of layer $l$ given input $x \in \mathcal D$ as follows:

\begin{definition}[Activation Pattern]
    Let $\mathcal N_{\theta}(x_0)$ be the application of a FNN $\mathcal N$ with parameters $\theta$ on an input $x_0 \in \mathcal D$, with $\mathcal D \subseteq \mathbb R^{n_0}$. Then, by referring to $x_{l-1}$ as the input to layer $l \in \{1,\dots, L\}$, we can compute the activation pattern $A_l(x_0)$ of layer $l$ on input $x_0$ as follows:
  \begin{equation}
    A_l(x_0) =
    \{ a_i \ | \ a_i = 1 \text{ if } h_{l,i}(x_{l-1}) > 0
    \text{ else } a_i = 0,
    \ \forall i = 1,\dots,n_l \}.
    \end{equation}
Thus, we can represent $ A_l(x_0)$ as a vector in $\{0,1\}^{n_l}$, i.e.:
\begin{equation}
     A_l(x_0) = [a_1,a_2,\dots,a_{n_l}].
\end{equation}

\end{definition}

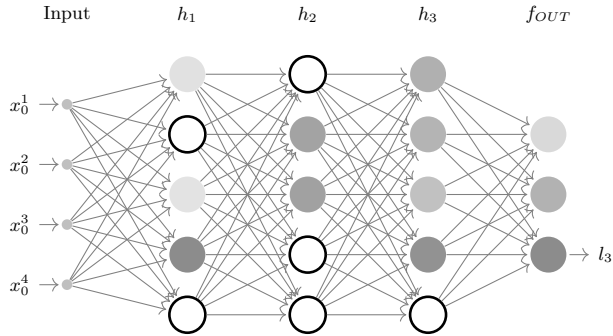
\begin{figure}[h]
    \centering
        
    \def\layersep{2cm}
    
    \begin{tikzpicture}[
        shorten >=1pt,
        ->,
        draw=black!50, 
        scale=0.8, every node/.style={scale=0.8},
        node distance=\layersep]
        \tikzstyle{every pin edge}=[<-,shorten <=1pt]
        \tikzstyle{neuron}=[circle,fill=black!25,minimum size=17pt,inner sep=0pt]
        \tikzstyle{input neuron}=[neuron,minimum size=5pt];
        \tikzstyle{output neuron}=[neuron, fill=gray];
        \tikzstyle{hidden neuron}=[neuron, fill=gray];
        \tikzstyle{annot} = [text width=4em, text centered]

        \foreach \name / \y in {1,...,4}
            \node[input neuron, pin=left:$x_0^\y$] (I-\name) at (0,-\y) {};

        \foreach \name / \y in {2,5} {
            \path[yshift=0.5cm]
                node[hidden neuron, fill = white, line width=1pt, draw=black] (H1-\name) at (\layersep,-\y cm) {};
        }
        
        \foreach \name / \y in {1,3,4} {
            \path[yshift=0.5cm]
                node[hidden neuron, opacity={max({(rand+1)/2},.1)}] 
                (H1-\name) at (\layersep,-\y cm) {};
        }

        \foreach \name / \y in {1,4,5} {
            \path[yshift=0.5cm]
                node[hidden neuron, fill = white, line width=1pt, draw=black] 
                (H2-\name) at (2*\layersep,-\y cm) {};
        }
        
        \foreach \name / \y in {2,3} {
            \path[yshift=0.5cm]
                node[hidden neuron, opacity={max({(rand+1)/2},.1)}] 
                (H2-\name) at (2*\layersep,-\y cm) {};
        }
        
        \foreach \name / \y in {5} {
            \path[yshift=0.5cm]
                node[hidden neuron, fill = white, line width=1pt, draw=black] 
                (H3-\name) at (3*\layersep,-\y cm) {};
        }

        \foreach \name / \y in {1,2,3,4}
            \path[yshift=0.5 cm]
                node[hidden neuron, opacity={max({(rand+1)/2},.1)}] 
                (H3-\name) at (3*\layersep,-\y cm) {};
        
        \foreach \name / \y in {1,...,2}
            \path[yshift=-0.5 cm]
                node[output neuron, opacity=\y*0.3] 
                (O-\name) at (4*\layersep,-\y cm) {};
        
        \foreach \name / \y in {3,...,3}
            \path[yshift=-0.5 cm]
                node[output neuron, opacity=\y*0.3, pin={[pin edge={->}]right:$l_{\y}$}] 
                (O-\name) at (4*\layersep,-\y cm) {};
                
        \foreach \source in {1,...,4}
            \foreach \dest in {1,...,5}
                \path (I-\source) edge (H1-\dest);
        
        \foreach \source in {1,...,5}
            \foreach \dest in {1,...,5}
                \path (H1-\source) edge (H2-\dest);
        
        \foreach \source in {1,...,5}
            \foreach \dest in {1,...,5}
                \path (H2-\source) edge (H3-\dest);
    
        \foreach \source in {1,...,5}
            \foreach \dest in {1,...,3}
                \path (H3-\source) edge (O-\dest);

        \node[annot,above of=H1-1, node distance=1cm] (h1l) {$h_1$};
        \node[annot,left of=h1l] {Input};
        \node[annot,right of=h1l] (h2l) {$h_2$};
        \node[annot,right of=h2l] (h3l) {$h_3$};
        \node[annot,right of=h3l] {$f_{OUT}$};
    \end{tikzpicture}
    \caption{The evaluation of the example neural network $\mathcal N$ on an input instance $ x_0 = [x_0^1, x_0^2, x_0^3, x_0^4]$. Hidden units have different opacity depending on the module of the positive output, while the black border indicates output 0.  In the last layer, the output label $l_3$ indicates the output unit with the largest value.
    In this example, we have: $A_1 = [1,0,1,1,0]$, $A_2 = [0,1,1,0,0]$, $A_3 = [1,1,1,1,0]$.}
    \label{fig:nn_eval}
\end{figure}

In Fig. \ref{fig:nn_eval} we show a simple example of a FNN $\mathcal N(x_0)$ and its activation patterns. In the following, we will represent generic activation patterns as $a$ or $a_i$, and with  $\mathtt{layer}(a)$ we will refer to the layer corresponding to that pattern. In addition, we allow us to simplify the notation of $A_l$ and refer to $A_l(\mathcal X_0)$ on $\mathcal X_0  \subseteq \mathbb R^{n_0}$ as $A_l(\mathcal X_0) = \bigcup_{x_0 \in \mathcal X_0} A_l(x_0)$. 

Given an activation pattern $\hat a$, or a set of patterns $\mathcal A$ belonging to different layers, and a set of instances $\mathcal X \subseteq \mathcal D$, we call \emph{activation region} the set composed by the instances in $\mathcal X$ that generate that activation pattern, or patterns, in their respective layers.

\begin{definition}[Activation Region]
    The activation region identified by an activation pattern $\hat a$ on an input subset $\mathcal X \subseteq \mathcal D$ is given by:
    
    \begin{equation}
    \mathcal{AR}(\hat a, \mathcal X) = \{ x \in \mathcal X \ | \ A_l(x) = \hat a, \ l = \mathtt{layer}(\hat a)\}.           
    \end{equation} 
    Given a set of activation patterns $\mathcal A$ belonging to different layers, i.e., $\forall a_i,a_j \in \mathcal A$ $\mathtt{layer}(a_i) \neq \mathtt{layer}(a_j)$, we define their activation region as:
    
    \begin{equation}
        \mathcal{AR}(\mathcal A, \mathcal X) = \bigcap_{\hat a \in \mathcal A}    \mathcal{AR}(\hat a, \mathcal X). 
    \end{equation}
\end{definition}

Given a dataset $\mathcal D$ and a network $\mathcal N_\theta$, we introduce the APD as the directed acyclic graph defined by all the activation patterns generated by instances in $\mathcal D$ and the way in which they are composed.

\begin{definition}[Activation Patterns DAG]
    Given a network $\mathcal N_\theta$ and a dataset $\mathcal D \subseteq \mathbb R^{n_0}$, the \emph{\textbf{A}ctivation \textbf{P}atterns \textbf{D}AG (\textbf{APD})} is a directed acyclic graph $APD_\mathcal {N_\theta}(\mathcal D) = (V, E)$, where:
    \begin{itemize}
        \item $V$ is the set of vertices defined by
        \[
        V = \{1, ..., |\mathcal A|\},
        \]
        where $\mathcal A = \bigcup_{l=1}^L A_l(\mathcal D)$ is the set of all possible activation patterns and $|\mathcal A|$ is its cardinality. In addition, let $\mathtt{patt}: V \rightarrow \mathcal A$ be a labelling function that associates each vertex to the corresponding activation pattern.
        \item $E$ is the set of edges defined by:
        \begin{equation}
            E = \{ (v_1, v_2) \in V \times V \ | \ \mathtt{patt}(v_1) \text{ and } \mathtt{patt}(v_2) \text{ are consecutive} \}, \end{equation} 
        where two patterns $a_i, a_j$ are called consecutive if \[
        \mathtt{layer}(a_i)
        = l
        = \mathtt{layer}(a_j) - 1
        \]
        and exists $x \in \mathcal D$ such that $A_l(x) = a_i$ and $A_{l+1}(x) = a_j$.
    \end{itemize}
\end{definition}

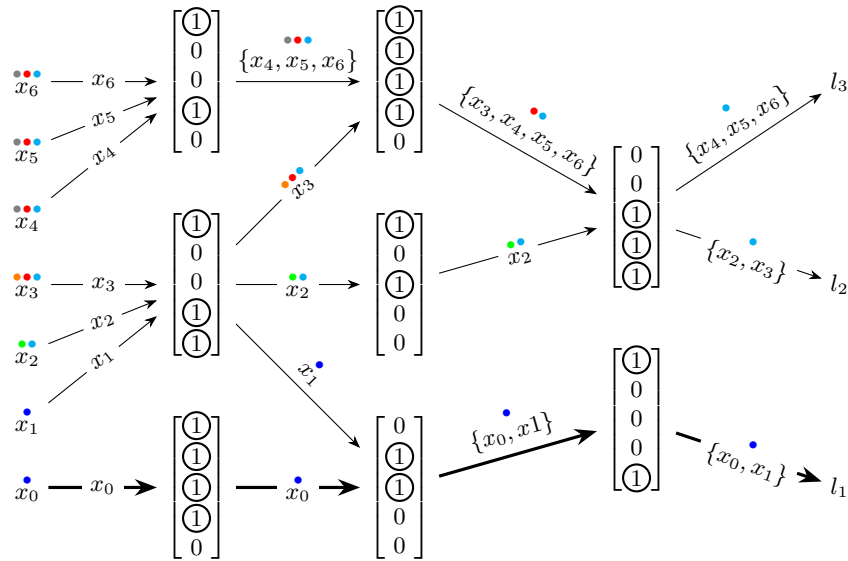
\begin{figure}[ht]
    \centering
        
    \newcommand{\onebl}{
    \tikz[baseline=(char.base)]{
            \node[shape=circle,draw=black,inner sep=1pt,thick] (char) {1};}
    }
    \newcommand{\bull}[1]{
        \textcolor{#1}{\bullet}
    }
    \newcommand{\clust}[2]{
        \ensuremath{\overset{#2}{#1}}
    }

    \begin{tikzpicture}[
        scale=0.9, every node/.style={scale=0.7}]
        \begin{scope}[every node/.style={}]
            \node (I1) at (-3,0) 
                {$\clust{x_0}{\bull{blue}}$};
            \node (I2) at (-3,1) 
                {$\clust{x_1}{\bull{blue}}$};
            \node (I3) at (-3,2) 
                {$\clust{x_2}{\bull{green}\bull{cyan}}$};
            \node (I4) at (-3,3) 
                {$\clust{x_3}{\bull{orange}\bull{red}\bull{cyan}}$};
            \node (I5) at (-3,4) 
                {$\clust{x_4}{\bull{gray}\bull{red}\bull{cyan}}$};
            \node (I6) at (-3,5) 
                {$\clust{x_5}{\bull{gray}\bull{red}\bull{cyan}}$};
            \node (I7) at (-3,6) 
                {$\clust{x_6}{\bull{gray}\bull{red}\bull{cyan}}$};
            \node (L11) at (-.5,0) 
                {$\begin{bmatrix}\onebl\\\onebl\\\onebl\\\onebl\\0\end{bmatrix}$};
            \node (L12) at (-.5,3)
                {$\begin{bmatrix}\onebl\\0\\0\\\onebl\\\onebl\end{bmatrix}$};
            \node (L13) at (-.5,6)
                {$\begin{bmatrix}\onebl\\0\\0\\\onebl\\0\end{bmatrix}$};
            \node (L21) at (2.5,0) 
                {$\begin{bmatrix}0\\\onebl\\\onebl\\0\\0\end{bmatrix}$};
            \node (L22) at (2.5,3)
                {$\begin{bmatrix}\onebl\\0\\\onebl\\0\\0\end{bmatrix}$};
            \node (L23) at (2.5,6)
                {$\begin{bmatrix}\onebl\\\onebl\\\onebl\\\onebl\\0\end{bmatrix}$};
            \node (L31) at (6,1)
                {$\begin{bmatrix}\onebl\\0\\0\\0\\\onebl\end{bmatrix}$};
            \node (L32) at (6,4)
                {$\begin{bmatrix}0\\0\\\onebl\\\onebl\\\onebl\end{bmatrix}$};
            \node (O3) at (9,6)
                {$l_3$};
            \node (O2) at (9,3)
                {$l_2$};
            \node (O1) at (9,0)
                {$l_1$};
        \end{scope}
    
        \begin{scope}[>={Stealth[black]},
                      every edge/.style={draw=black},
                      every node/.style={above,sloped},
                      innode/.style={fill=white, anchor=center, pos=0.5}]
            \path [->,very thick] (I1) edge node[innode] {$x_0$} (L11);
            \path [->] (I2) edge node[innode] {$x_1$} (L12);
            \path [->] (I3) edge node[innode] {$x_2$} (L12);
            \path [->] (I4) edge node[innode] {$x_3$} (L12);
            \path [->] (I5) edge node[innode] {$x_4$} (L13);
            \path [->] (I6) edge node[innode] {$x_5$} (L13);
            \path [->] (I7) edge node[innode] {$x_6$} (L13);
            
            \path [->,very thick] (L11) edge node[innode] {$\clust{x_0}{\bull{blue}}$} (L21);
            \path [->] (L12) edge node[innode] {$\clust{x_3}{\bull{orange}\bull{red}\bull{cyan}}$} (L23);
            \path [->] (L12) edge node[innode] {$\clust{x_2}{\bull{green}\bull{cyan}}$} (L22);
            \path [->] (L12) edge node {$\clust{x_1}{\bull{blue}}$} (L21);
            \path [->] (L13) edge node {$\clust{\{x_4, x_5, x_6\}}{\bull{gray}\bull{red}\bull{cyan}}$} (L23);
            
            \path [->,very thick] (L21) edge node {$\clust{\{x_0,x1\}}{\bull{blue}}$} (L31);
            \path [->] (L22) edge node[innode] {$\clust{x_2}{\bull{green}\bull{cyan}}$} (L32);
            \path [->] (L23) edge node {$\clust{\{x_3,x_4,x_5,x_6\}}{\bull{red}\bull{cyan}}$} (L32);
            
            \path [->,very thick] (L31) edge node[innode] {$\clust{\{x_0,x_1\}}{\bull{blue}}$} (O1);
            \path [->] (L32) edge node[innode] 
                {$\clust{\{x_2,x_3\}}{\bull{cyan}}$} (O2);
            \path [->] (L32) edge node {$\clust{\{x_4,x_5,x_6\}}{\bull{cyan}}$} (O3);
        \end{scope}
    \end{tikzpicture}
    \caption{The representation of $APD_{\mathcal N}(\{x_0, \dots, x_6\})$, where $x_0$ is the input instance evaluated in Fig. \ref{fig:nn_eval}. The thicker lines correspond to the edges generated by example $x_0$. In addition, on the right we represented the label predicted by the network. Finally, colored bullets over instances mark the splitting history (see Algo. \ref{algo:decision}), from right to left, i.e., each new bullet identifies a new partition of the previous cluster.}
    \label{fig:apd}
\end{figure}

In Fig. \ref{fig:apd} we show the $APD_\mathcal N$ defined by the network $\mathcal N$ of Fig. \ref{fig:nn_eval}, as generated on five example samples.

\subsection{Clustering the input dataset using the APD}
As discussed in the previous sections, a given activation pattern defines a specific activation region in the input space, to which one or more instances are associated. In fact, the same linear transformation can be shared by multiple instances, as it was analytically showed in \cite{pascanu_number_2014,montufar_number_2014}. 
For example, the linear transformation defined by pattern $a = [1,0,0,0,1]$ of the third layer in Fig. \ref{fig:apd} is common to both $x_0$ and $x_1$.  

Similarly to \cite{tao_deep_2019}, we here exploit the compositional structure of the the APD to characterize each input instance on the basis of the trajectory through  activation patterns in the distinct layers. 
The intuition is that the overlap among the trajectories of two instances on the APD is effective in assessing how the two instances are similarly processed throughout the network. 

Additionally, from Fig. \ref{fig:apd} one can notice that some activation patterns are characterized by a decision boundary, such as activation pattern $[0,0,1,1,1]$ of the third layer, whereas some are not, such as activation pattern $[1,0,0,1,0]$ of the first layer. We will refer to the former as \emph{unstable activation patterns} and to the latter as \emph{stable}. 
In this respect, it would be interesting to test whether instances belonging to stable activation patterns are the ones on which the network is more confident. 

Furthermore, in order to assess the similarity of the instances with respect to classification labels, it might be effective to look for class-specific stable activation patterns belonging to the last layers. The motivation is that we are interested in understanding which previous transformation has brought the instances close in the feature space, as a result of space folding, and if transformations are related with the predicted label. 
For example, in Figure \ref{fig:nn_eval} instances $x_4, x_5, x_6$ activate the same activation patterns from the beginning, while instances $x_1$ and $x_0$ are folded in the same activation pattern after the first layer transformation.

\begin{algorithm}[t]
    \centering
    
  \newcommand*\Let[2]{\State #1 $\gets$ #2}
  
  \begin{algorithmic}[0]
    \Function{split}{APD $G = (V, E)$, dataset $\mathcal{D}$, FNN $\mathcal N$}
      \Let{$n.\mathtt{pred()}$}{predecessors of node $n \in V$}
      \Let{L}{\# layers of $\mathcal N$}
      \Let{$out$}{dummy ending node}
      \For{$v \in V$ s.t. $\mathtt{layer}(\mathtt{patt}(v))==L$}
        \State $E.\mathtt{add}((v,out))$
      \EndFor
      \Let{$\mathcal P$}{$\{(out, \mathcal D) \})$} \Comment{Current partition}
      \Let{$\mathcal F$}{$\emptyset$} \Comment{Final partition}
      \While{$\mathcal P \neq \emptyset$}
      
        \Let{(n, $\mathcal{C}$)}{$\mathcal P.\texttt{pop}()$} \Comment{Extract (current node, cluster)}
        \If{$n.\mathtt{pred}() == \emptyset \lor |\mathcal C| == 1$} \Comment{Check if splittable cluster}
            \State $\mathcal F.\texttt{add}(\mathcal C)$
            \State \texttt{break}
        \EndIf
        \Let{$\mathcal S$}{$\emptyset$}
        \For{$v \in n.\mathtt{pred}()$} \Comment{Split current cluster}
            \Let{$\mathcal V$}{$\mathcal{AR}(\texttt{patt}(v),\mathcal C)$}
            \State $\mathcal S.\texttt{add}((v,\mathcal V))$
        \EndFor
        \Let{$\mathcal S'$}{$\{\mathcal V \ | \ (v,\mathcal V) \in \mathcal S\}$}
        \Let{$ig$}{\texttt{InformationGain}$(\mathcal C, \mathcal S')$}
        \If{$ig > 0$} \Comment{Check splitting gain}
          \Let{$\mathcal P$}{$\mathcal P \cup \mathcal S$}
        \Else
          \State $\mathcal F.\texttt{add}(\mathcal C)$
        \EndIf
      \EndWhile

      \State \Return{$\mathcal{F}$}
    \EndFunction
  \end{algorithmic}

    \caption{Splitting algorithm.}
    \label{algo:decision}
\end{algorithm}

To automatically identify similar instances, we defined a splitting algorithm, formally defined in Algo. \ref{algo:decision}. 
The goal is to cluster instances that share the same activation patterns and are classified with the same label, proceeding backwards from the bottom of the network. 
The first partition of input data is performed by considering only the activation patterns of the last layer; if one of the identified clusters contains instances with distinct labels, it is splitted by considering which activation pattern they activate in the previous layer. Splitting is determined via information gain measure \cite{quinlan_induction_1986}, since a decrease of entropy implies more homogeneous partitions. 

In Fig. \ref{fig:apd} colored bullets mark the splitting history of the $6$ instances. For example, the first partition is identified by cyan and blue color, i.e. $\{\{x_2,x_3,x_4,x_5,$ $x_6\},\{x_0,x_1\}\}$. Cluster $\{x_0, x_1\}$ is not splitted, because both instances are classified with $l_1$ label. Conversely, the other cluster is partitioned twice: the first splitting occurs when considering the second layer, as $x_2$ has a different activation pattern than the others and is classified with a different class; the same occurs at the first layer, this time between $x_3$ and the other instances. 
The final partition is the following $\{\{x_0,x_1\},\{x_4,x_5,x_6\},\{x_3\},\{x_2\}\}$.

In the next section, we will present some preliminary results on how cluster size of the instances partition can be used to evaluate input similarity and hardness.

\FloatBarrier
\section{Results}
\label{sec:results}

We applied the clustering algorithm discussed in the previous section on the MNIST dataset and tested it on ReLU networks with different architectures. 
In particular, we will show that instances included in largest clusters may be \enquote{easier} for the network, while errors and \enquote{hard} instances are usually included in small clusters. 
More in detail, we are looking for similar instances in the feature space that are classified with the same label, as this may be interpreted as a measure of \enquote{confidence} of the network in that specific composition of transformations. 

The experiments were performed with a fixed learning rate of $0.0001$, $500$ epochs, SGD as optimization algorithm and the following different architectures:
\begin{enumerate*}[label=(\roman*)]
    \item \texttt{32full}: $5$ layers with $32$ neurons each; \item \texttt{16full}: $5$ layers with $16$ neurons each; \item \texttt{32bottl}: with $5$ layers with $32, 16, 12, 10, 8$ neurons each.
\end{enumerate*} 
The accuracy obtained on the MNIST dataset were, respectively: $98.3 \%$ for \texttt{32full}, $97.2\%$ for \texttt{32bottl} and $95.8\%$ for \texttt{16full}.

\begin{figure}
    \centering
    \includegraphics[scale=.25]{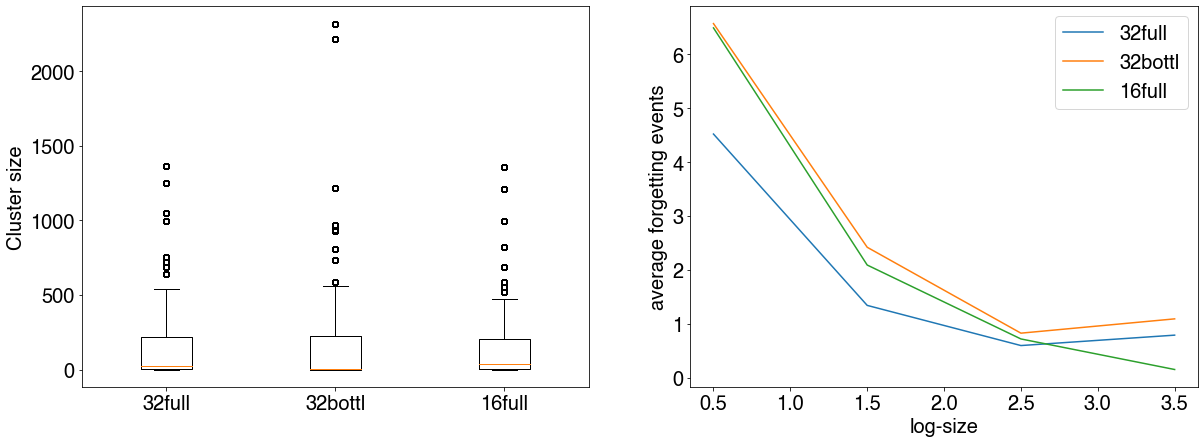}
    \caption{(Left) Boxplots of cluster size distributions with different architectures. (Right) Average number of forgetting events against log-binned cluster size.}
    \label{fig:box_avg}
\end{figure}

\begin{figure}
    \centering
    \includegraphics[trim={1.5cm 0 0 0},clip,scale=.34]{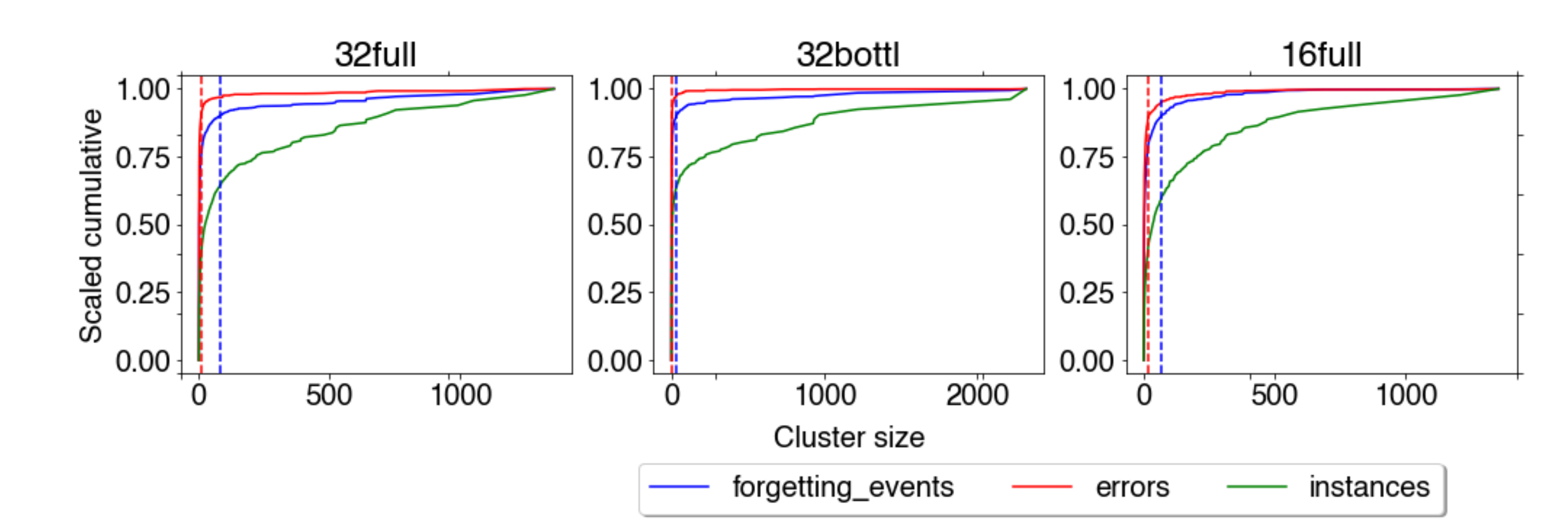}
    \caption{Cumulative distribution of forgetting events and errors by (sorted) cluster size on three different architectures. In green the cumulative number of considered instances. All the three lines are normalized between 0 and 1. The vertical dotted lines represent where $90\%$ of the respective cumulative is reached.}
    \label{fig:cum_forg}
\end{figure}

In Fig. \ref{fig:box_avg} (left) the distribution of the sizes of the input partition obtained for different architectures is reported. The majority of the clusters are small (average size $\approx 4, 3, 5$ for \texttt{32full}, \texttt{16full} and \texttt{32bottl}, respectively), while even very large clusters (containing up to $2000$ instances) are observed for all architectures. 
Bigger clusters are expected to contain a larger number of correctly classified instances, i.e., the instances on which the network is more \enquote{confident}. To test our hypothesis, we analyzed the distribution of forgetting events by cluster size, where forgetting events are defined as follows:
\begin{definition}[Forgetting event \cite{toneva_empirical_2018}]
    \label{def:forg_event}
    Let $x$ be an instance with label $k$ and $pred_e(x)$ the predicted label of $x$ at epoch $e$. A learning event at epoch $e$ occurs when $pred_{e-1} \neq k$ and $pred_{e} = k$. A forgetting event at epoch $e$ occurs when $pred_{e-1} = k$ and $pred_{e} \neq k$. If an instance has no forgetting event during the learning process, is called unforgettable, otherwise is a forgettable instance.
\end{definition}
In Fig. \ref{fig:box_avg} (right) we display the average number of forgetting events with respect to (log-binned) cluster size. From the picture it seems to emerge that, for all architectures, the forgettable instances are grouped in the small clusters. This trend is confirmed by looking at the cumulative distributions of errors and forgetting events in Fig. \ref{fig:cum_forg}.

\begin{figure}
        \begin{subfigure}[t]{0.25\textwidth}
            \centering
            \includegraphics[trim={.4cm 0 2cm 0},clip,scale=.19]{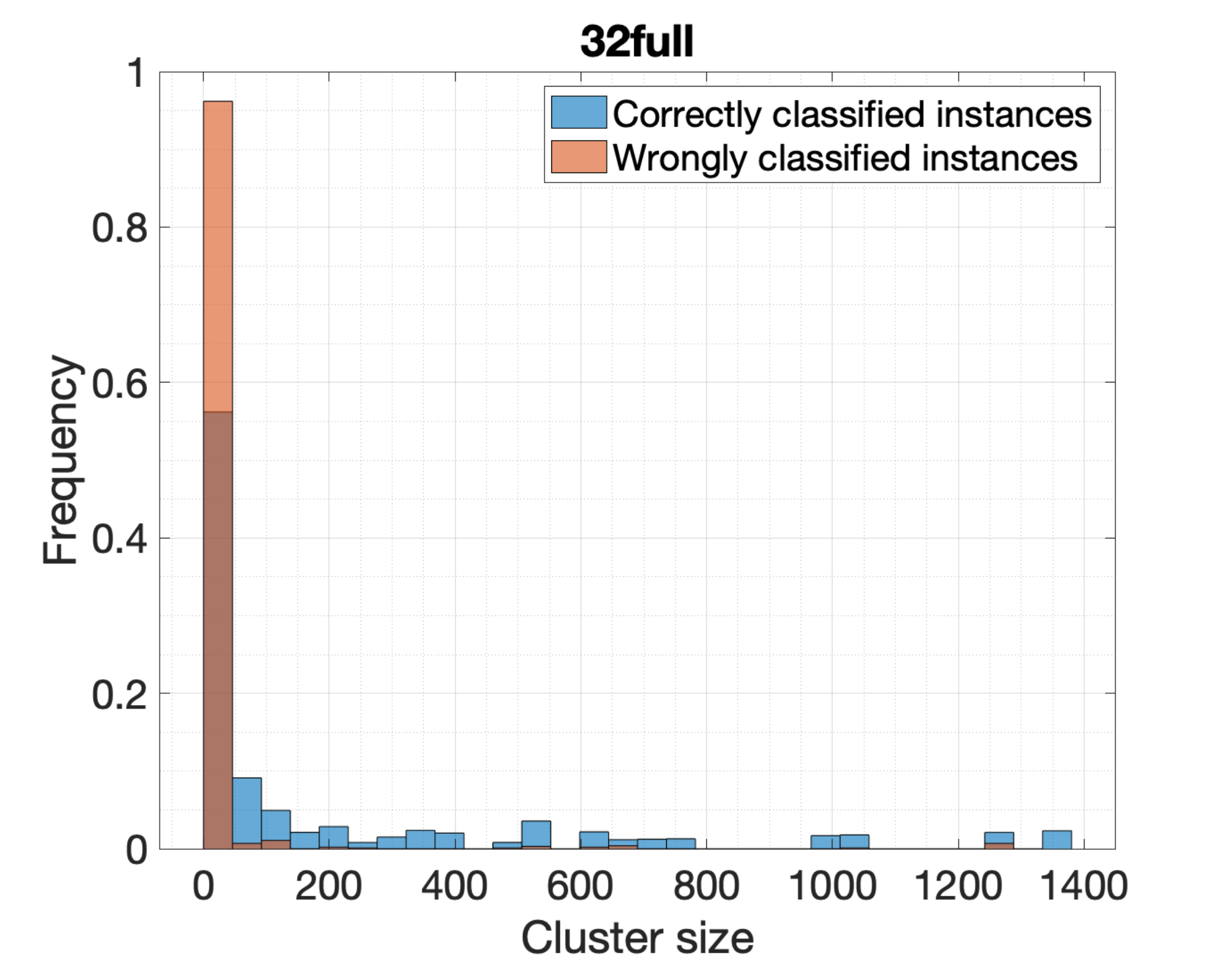}
        \end{subfigure}
        \hfill
        \begin{subfigure}[t]{0.25\textwidth}
            \centering
           \includegraphics[trim={.4cm 0 2cm 0},clip,scale=.19]{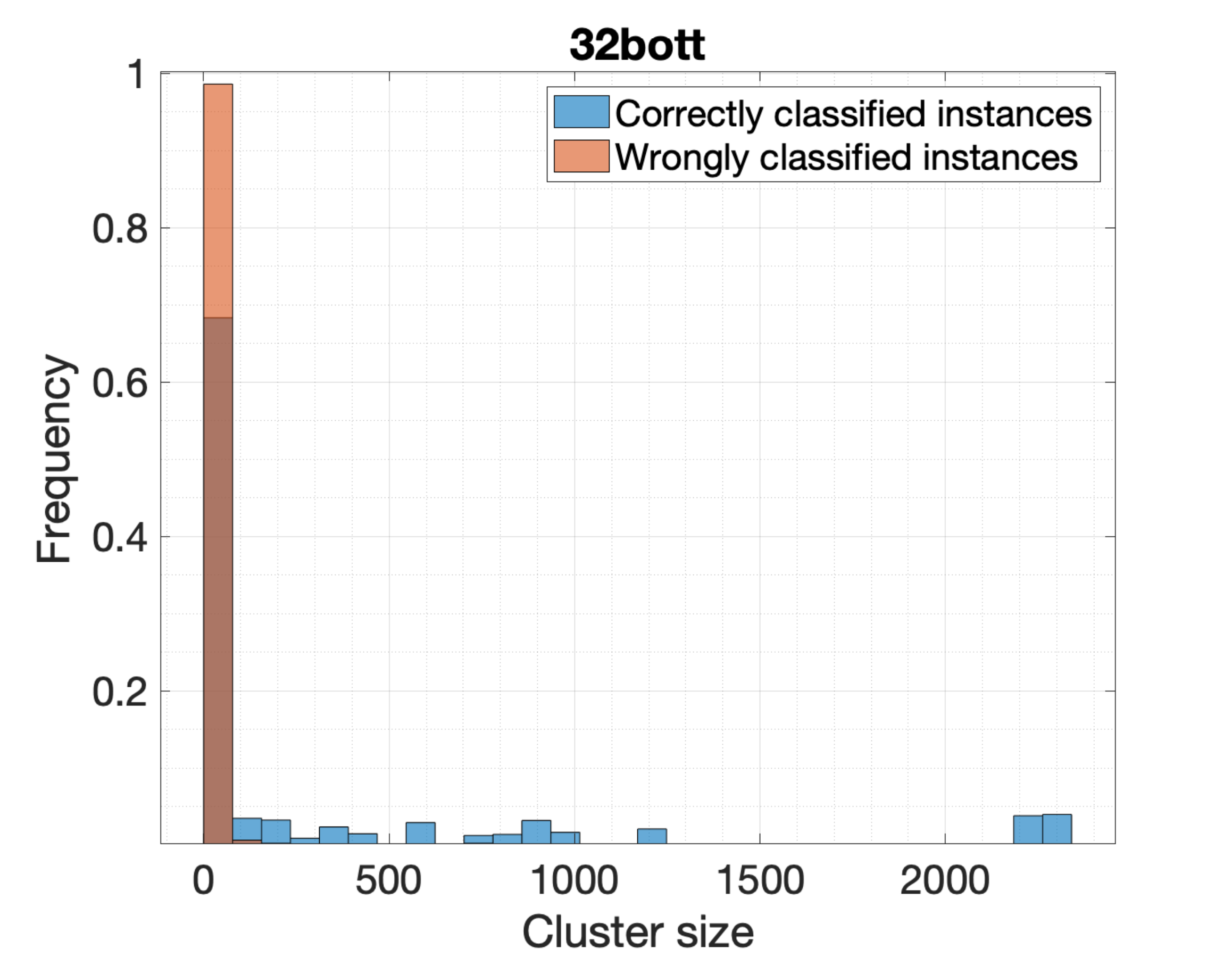}
        \end{subfigure}
        \hfill
        \begin{subfigure}[t]{0.33\textwidth}
            \includegraphics[trim={.4cm 0 2cm 0},clip,scale=.19]{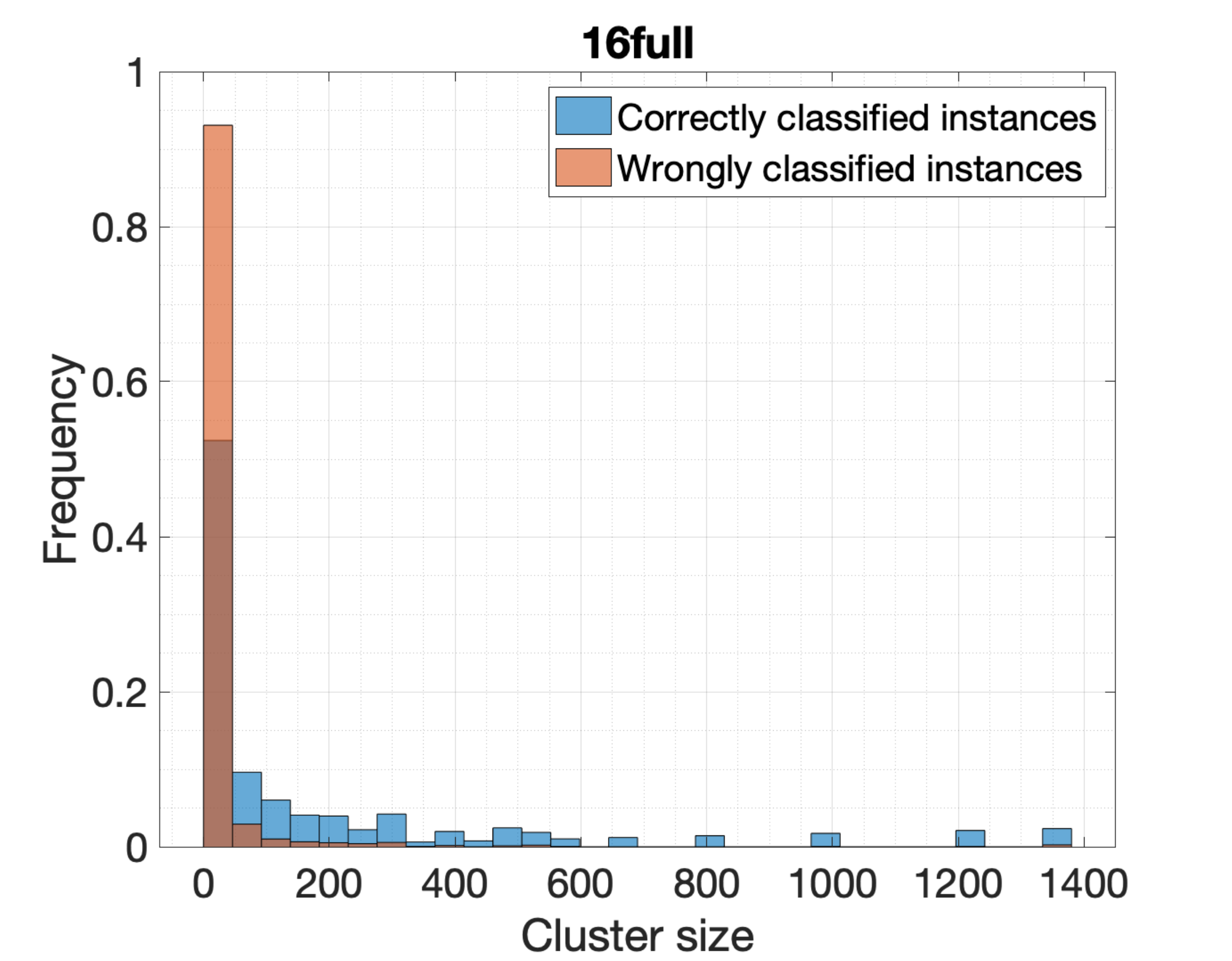}
        \end{subfigure}
        \caption{Cluster size distribution for correctly and wrongly classified instances.}
        \label{fig:hist}
    \end{figure}

Finally, in Fig. \ref{fig:hist} one can see the distribution of the cluster size with respect to either correctly and wrongly classified instances. 
Consistently with the other findings, wrongly classified instances are characterized by very small clusters (mostly singletons) for all architectures, whereas correctly classified instances are typically included in clusters with significantly larger size and a much higher variance. 
Again, this result would suggests the presence of a significant correlation between cluster size and the input hardness. 

\section{Conclusions and future developments}

In this work we introduced the APD, a structure that represents the composition of piecewise linear functions defined by the layers of a ReLU network. 
Additionally, we proposed an algorithm to partition the input dataset based both on the composition of linear transformations defined by the layers and the predicted labels. 
We showed that this partition can be efficiently used to group the instances that are similarly transformed by network. Furthermore, we speculate that the instances included in large clusters are those on which the network is more confident and that are better classified, an hypothesis that was confirmed by the preliminary tests performed on the MNIST dataset.

This new framework might be extremely useful to improve the interpretation of the inner representation of DNNs, and could be extended as follows.

 \emph{Intrinsic Dataset Dimension}: one could estimate the intrinsic dataset dimensionality by considering the distribution of the inputs after applying our clustering algorithm. The idea is that a dataset with many similar (redundant) instances has a smaller dimension than a dataset where all instances are singletons. The same analysis was performed in \cite{toneva_empirical_2018}, by using the number of forgettable instances as an estimate of the dataset dimensionality.
    
 \emph{Dataset Reduction}: one could use our clustering method to discriminate between overfitted instances (i.e., with a few similar instances) and \enquote{easy} instances. By iteratively training a new network only on the overfitted instances of the previous one, one could build an ensamble of DNNs with the aim of increasing accuracy, as similarly proposed in \cite{tao_deep_2019}. 

 \emph{Sample weighting}: one could use the input partition as a sample weighting technique during learning, such as \emph{self-paced learning} \cite{kumar_self-paced_2010} or \emph{hardness mining} \cite{katharopoulos_not_2018}.  

Clearly, these results were obtained on a small selection of the possible contributing factors, therefore we will extend our analysis by considering other conditions, such as different optimization algorithms or network architectures. In particular, we are going to consider other types of input data, since computer vision datasets might induce a bias in our analysis due to their specific structure.

In conclusion, the APD represents a simple, but expressive tool, to study how DNNs learn data, motivated by geometrical studies on the properties of PWL activation functions \cite{montufar_number_2014,pascanu_number_2014}.

\FloatBarrier

\emergencystretch=2em
\printbibliography
\end{document}